\documentclass[
]{ceurart}

\begin{document}


\conference{}

\title{Fit to Measure: Reasoning about Sizes for Robust Object Recognition}

\author[1]{Agnese Chiatti}[%
orcid=0000-0003-3594-731X,
]
\ead{agnese.chiatti@open.ac.uk}
\address[1]{Knowledge Media Institute, The Open University,
  Walton Hall, Milton Keynes, MK7 6AA, United Kingdom}

\author[1]{Enrico Motta}[%
orcid=0000-0003-0015-1592,
]
\ead{enrico.motta@open.ac.uk}

\author[1]{Enrico Daga}[%
orcid= 0000-0002-3184-540,
]
\ead{enrico.daga@open.ac.uk}

\author[1]{Gianluca Bardaro}[%
orcid=0000-0002-6695-0012,
]
\ead{gianluca.bardaro@open.ac.uk}

\begin{abstract}
  Service robots can help with many of our daily tasks, especially in those cases where it is inconvenient or unsafe for us to intervene – e.g., under extreme weather conditions or when social distance needs to be maintained. However, before we can successfully delegate complex tasks to robots, we need to enhance their ability to make sense of dynamic, real-world environments. In this context, the first prerequisite to improving the Visual Intelligence of a robot is building robust and reliable object recognition systems. While object recognition solutions are traditionally based on Machine Learning methods, augmenting them with knowledge-based reasoners has been shown to improve their performance. In particular, based on our prior work on identifying the epistemic requirements of Visual Intelligence, we hypothesise that knowledge of the typical size of objects could significantly improve the accuracy of an object recognition system. To verify this hypothesis, in this paper we present an approach to integrating knowledge about object sizes in a ML-based architecture. Our experiments in a real-world robotic scenario show that this combined approach ensures a significant performance increase over state-of-the-art Machine Learning methods.
\end{abstract}


\maketitle

\section{Introduction}

With the fast-paced advancement of the Artificial Intelligence (AI) and Robotics fields, there is an increasing potential to resort to\textit{ robot assistants} (or \textit{service robots}) to help with daily tasks. Service robots can take on many roles. They can operate as patient carers \cite{bajones_hobbit_2018}, door-to-door garbage collectors \cite{ferri_dustcart_2011}, Health and Safety monitors \cite{dong_design_2018}, museum or tour guides \cite{waldhart_reasoning_2019}, to name a few. Yet, even decades after the first robot vacuum cleaner was deployed (\url{https://en.wikipedia.org/wiki/Robotic_vacuum_cleaner}), robot assistants still perform unreliably on more complex tasks. Succeeding in the real world is indeed a difficult challenge because it requires robots to make sense of the high-volume and diverse data coming through their perceptual sensors. Although different sensory modalities contribute to the robot’s \textit{sensemaking} abilities (e.g., touch, sound, temperature), in this work, we focus on the modality of vision. From this entry point, the problem then becomes one of enabling robots to correctly interpret the stimuli of their vision system, with the support of background knowledge sources, a capability also known as \textit{Visual Intelligence} \cite{chiatti_towards_2020}. The first prerequisite to Visual Intelligence is the ability to robustly recognise the different objects occupying the robot’s environment (\textit{object recognition}). Let us consider the case of HanS, the Health and Safety robot inspector currently under development at the Knowledge Media Institute (KMi). HanS is expected to monitor the Lab space in search of potentially dangerous situations, such as fire hazards. For instance, imagine that Hans was observing a flammable object (e.g., a paper cup) left on top of a portable heater. To conclude that it is in the presence of a potential fire hazard, the robot would first need to recognise that the cup and the portable heater are there. 

Currently, the most common approach to tackling object recognition tasks is applying methods which are based on Machine Learning (ML). In particular, the state-of-the-art performance is defined by the latest approaches based on Deep Learning (DL) \cite{liu_deep_2020,lecun_deep_2015}. Despite their popularity, these methods have received many critiques due to their brittleness and lack of transparency \cite{marcus2018deep,parisi_continual_2019,pearl_theoretical_2018}. To compensate for these limitations, a more recent trend among AI researchers has been to combine ML with knowledge-based reasoning, thus adopting a \textit{hybrid approach} \cite{aditya_integrating_2019,gouidis_review_2020}. A question remains, however, on what type of knowledge resources and reasoning capabilities should be leveraged within this new class of hybrid methods \cite{daruna2018sirok}.

In \cite{chiatti_towards_2020}, we identified a set of \textit{epistemic requirements}, i.e., a set of capabilities and knowledge properties, required for service robots to exhibit Visual Intelligence. We then mapped the identified requirements to the types of classification errors emerging from one of HanS’ scouting rounds, where we relied solely on Machine Learning to recognise the objects. This error analysis highlighted that, in 74\% of the cases, a more accurate object classification could in principle have been achieved if the relative size of objects was considered for their categorisation. For instance, back to HanS’ case, the paper cup could be mistaken for a rubbish bin, due to its shape. However, rubbish bins are typically larger than cups. With this awareness, HanS would be able to rule out object categories, which, albeit being visual similarity to the correct class, are \textit{implausible} from the standpoint of size.

These elements of \textit{typicality} and \textit{plausible reasoning} \cite{davis_commonsense_2015} link size reasoning to the broader AI objective of developing systems which exhibit \textit{common sense} \cite{levesque_common_nodate}, especially with respect to understanding a set of intuitive physics rules governing the environment \cite{hayes_second_1988,lake_building_2017}. This view is also supported by studies of human visual cognition, which suggest that our priors about the canonical size of objects play a role in how we categorise, draw and imagine objects \cite{rosch_principles_1999,hoffman_visual_2000,konkle_canonical_2011}.

On a more practical level, knowledge representations which encode object sizes have already been applied effectively to Natural Language Processing (NLP) tasks. These include answering questions, such as, “is object A larger than object B?” \cite{bagherinezhad_are_2016,elazar_how_2019}. However, despite this body of theoretical and empirical evidence, the role of size in object recognition has received little attention in the field of Computer Vision. To address this issue, in this paper we investigate the performance effects of augmenting a ML-based object recognition system both with background knowledge about the typical size of objects, as well as with a method to reason about the size of the observed objects. Namely, we propose: 

\begin{itemize}
\item A hybrid method to validate ML-based predictions based on the typical size of objects. 
\item A novel representation for size, which categorises objects differently along different dimensions contributing to their size (i.e., their front surface area, depth and aspect ratio). This representation also allows us to model object categories that include instances of varying size.
\end{itemize}

\section{Related work}

State-of-the-art object recognition methods rely heavily on Machine Learning, as further discussed in Section 2.1. Because of the limitations of ML-based methods, hybrid methods, which combine ML with background knowledge and knowledge-based reasoning, have been recently proposed (Section 2.2). In particular, our hypothesis is that awareness of object size has the potential to drastically improve the performance of hybrid object recognition methods \cite{chiatti_towards_2020}. Therefore, we will conclude our review of the literature by discussing existing approaches to representing the size of objects (Section 2.3). 

\subsection{Machine Learning for Object Recognition}
The impressive performance exhibited by object recognition methods based on DL has led to significant advances on several Computer Vision benchmarks \cite{liu_deep_2020,krizhevsky_imagenet_2017,he_deep_2016}. Deep Neural Networks (NNs), however, come with their own limitations. These models (i) are notoriously data-hungry, i.e., require thousands of annotated training examples to learn from, (ii) learn classification tasks offline, i.e., assume to operate in a closed world, on a fixed and pre-determined set of objects \cite{mancini_knowledge_2019}, and (iii) learn representational patterns automatically, by iterating over a raw input set \cite{lecun_deep_2015}. The latter trait can drastically reduce the start-up costs of feature engineering. However, it also complicates tasks such as explaining the obtained features and augmenting them with explicit knowledge statements \cite{marcus2018deep,pearl_theoretical_2018}.

The issue of learning robust object representations to adequately reflect changes in the environment, even from minimal training examples, has inspired the development of few-shot metric learning methods. \textit{Metric learning} is the task of learning an embedding (or feature vector) space, where similar objects are mapped closer to one another than dissimilar objects. In this setup, even objects unseen at training time can be categorised, by matching the learned representations against a support (reference) image set. In particular, in a \textit{few-shot scenario}, the number of training examples and support images is kept to a minimum. Deep metric learning has been applied successfully to object recognition tasks \cite{koch_siamese_2015,hoffer_deep_2015,schroff_facenet_2015}, even in real-world, robotic scenarios \cite{zeng2018robotic}. Koch and colleagues \cite{koch_siamese_2015} originally proposed to train two identical Convolutional Neural Networks (CNN) fed with images to be matched by similarity. This twin architecture is also known as Siamese Network. An extension of the Siamese architecture is the Triplet Network \cite{hoffer_deep_2015,schroff_facenet_2015}, where the input data are fed as triplets including: (i) one image depicting a certain object class (i.e., \textit{anchor}), (ii) a \textit{positive example} of the same object, and (ii) a \textit{negative example}, depicting a different object. The winning team for the object stowing task at the latest Amazon Robotic Challenge further tested the effects of learning weights independently on each CNN branch \cite{zeng2018robotic}. Relaxing the weight-coupling constraint of Siamese and Triplet Networks was shown to benefit the matching of images across different visual domains, e.g., robot-collected images with product catalogue images. Hence, in what follows, we will use the two top-performing solutions in \cite{zeng2018robotic} as a baseline to evaluate the object recognition performance of solutions which are purely based on Machine Learning. 

\subsection{Hybrid Methods for Object Recognition}
Broadly speaking, \textit{hybrid reasoning methods} combine knowledge-based reasoning with Machine Learning. A detailed survey of hybrid methods developed to interpret the content of images can be found in \cite{aditya_integrating_2019,gouidis_review_2020}. Many of these hybrid methods are specifically tailored on Deep NNs, which define the predominant approach to tackling object recognition problems. In this setup, background knowledge and knowledge-based reasoning can be integrated at four different levels of the NN \cite{aditya_integrating_2019}: (i) in \textbf{pre-processing}, to augment the training examples, (ii) within the \textbf{intermediate layers}, (iii) as part of the \textbf{architectural topology} or \textbf{optimisation function}, and (iv) in the \textbf{post-processing} stages, to validate the NN predictions.

Methods in the first group rely on external knowledge to compensate for the lack of training examples. In \cite{mancini_knowledge_2019}, auxiliary images depicting newly-encountered objects were first retrieved from the Web and then manually validated. As a result, significant supervision costs were introduced to compensate for the noisiness of data mined automatically.

Other approaches have encoded the background knowledge directly in the inner layer representations of a Deep NN. In the RoboCSE framework, a set of knowledge properties of objects, (i.e., their typical location, fabrication material and affordances) were represented through multi-relational embeddings \cite{daruna_robocse_2019}. This method was proven effective to infer the object’s material, location and affordances from its class, but performed poorly on object categorisation tasks (i.e., when asked to infer the object’s class from its properties).

More transparent and explainable than multi-relational embeddings, methods in the third group are either inspired by the topology of external knowledge graphs \cite{marino_more_2017} or introduce reasoning components which are trainable end-to-end \cite{serafini_logic_2016,manhaeve_deepproblog_2018,santoro_simple_2017,van_krieken_analyzing_2020}. Graph Search Neural Networks (GSNN) \cite{marino_more_2017} resemble an input knowledge graph, where search seeds are selected based on a separate object detection module. In Logic Tensor Networks (LTN) \cite{serafini_logic_2016}, entities are represented as distinct points in a vector space, based on a set of soft-logic assertions linking these entities. In this framework, symbolic rules (which may adhere to probabilistic logic - \cite{manhaeve_deepproblog_2018}) are added as constraints to the NN’s optimisation function. Authors in \cite{santoro_simple_2017} proposed to incorporate reasoning in the form of a trainable Relational Reasoning Layer. Similarly, in \cite{van_krieken_analyzing_2020}, differentiable knowledge statements (expressed in fuzzy logic) contribute to the training loss function, to aid digit classification on the MNIST dataset.

Finally, the fourth family of hybrid methods uses knowledge-based reasoning to validate the object predictions generated through ML. In \cite{young_towards_2016,young_semantic_2017} the results produced by the ML-based Vision module are first associated with the Qualitative Spatial Relationships extracted from the input image, and then also matched against the top-most related DBpedia concepts, if an unknown object is observed. As in the case of \cite{santoro_simple_2017,van_krieken_analyzing_2020}, methods in this group can modularly interface with different NN architectures. Moreover, they make it possible to reason on objects unseen at training time, by querying external knowledge sources. For these reasons, in the approach proposed in this paper, knowledge-based reasoning is applied after generating the ML predictions. Because our focus is on reasoning about size, we cannot directly compare our approach against methods in \cite{young_towards_2016,young_semantic_2017}, which focus on spatial reasoning and taxonomic reasoning (i.e., reasoning about the semantic relatedness of different object categories). 

\subsection{Representing the Size of Objects}
Studies on human visual cognition have suggested that there seems to be a set of preferred (or canonical) views which we use to mentally represent objects \cite{rosch_principles_1999,hoffer_deep_2015}. These views have recognisable colour and shape features. Similarly, our perception of the object’s sizes is influenced by a set of prototypical priors. Specifically, the \textit{canonical size} we use to imagine and draw a certain object appears to be proportionally related to the logarithm of the object’s assumed size, i.e., our “prior knowledge about the size of objects in the world” \cite{konkle_canonical_2011}. Inspired by these findings, Bagherinezad et al. \cite{bagherinezhad_are_2016} have modelled the object sizes through a log-normal distribution. The resulting distributions were then used to populate nodes in a graph, where objects which co-occurred frequently across the YFCC100M dataset \cite{thomee_yfcc100m_2016} were linked together.

The size representation proposed in \cite{bagherinezhad_are_2016} is both \textit{quantitative} (i.e., expressed through statistical descriptors) and \textit{qualitative }(i.e., smaller than or larger than, as symbolised by the graph’s edges). Another quantitative representation was proposed in \cite{elazar_how_2019}, where sizes (namely the object’s length or volume) are represented as Distributions over Quantities (DoQ). In \cite{zhu_reasoning_2014}, descriptors of the object’s length are instead quantised with respect to three qualitative bins (i.e., <10in, 10–100in and >100in). Compared to \cite{zhu_reasoning_2014}, the statistical distributions in \cite{bagherinezhad_are_2016,elazar_how_2019} can more expressively model the size variety within the same object class. Indeed, the same object class can comprise of many \textit{instantiations} or \textit{models} (a short novel and a dictionary are both books, although dictionaries are usually thicker). Moreover, the same object can be observed under different \textit{appearances} (e.g., the book could be open or closed). Therefore, relying on a knowledge representation which can capture this within-class variability is crucial to ensure reuse across different real-world scenarios.

All the three approaches rely on data retrieved from the Web; this approach significantly reduces the cost of hardcoding a Knowledge Base about object sizes, but it is also more sensitive to noise. In addition, a limitation of the reviewed representations is that size is represented in one-dimensional terms, e.g., either through the object’s volume or through its length. However, different physical dimensions (height, width and depth) contribute differently to characterising an object. For instance, recycling bins and coat stands may occupy a comparable volume, but bins are usually thicker than coat stands.

The size representations adopted in \cite{bagherinezhad_are_2016,elazar_how_2019} have enabled to answer questions posed in natural language, such as “are elephants bigger than butterflies?”. Furthermore, in \cite{zhu_reasoning_2014}, the size features were used, among others, as an intermediate representation to predict the objects’ \textit{affordances}, or typical uses. Nonetheless, the role of size in object recognition is yet to be evaluated. In this paper, we propose a novel qualitative representation for the object’s typical size, which requires minimal manual annotations and controls for the presence of noisy measurements.

\section{Methodology}

\subsection{Representing qualitative sizes in a Knowledge Base}
We identified 60 object categories which are commonly found in KMi, the setting in which we aim to deploy our robotic Health and Safety monitor, HanS. These include not only objects which are common to most office spaces (e.g., chairs, desktop computers, keyboards), but also Health and Safety equipment (e.g., fire extinguishers, emergency exit signs, fire assembly point signs), and objects which are, to some extent, specific to KMi (e.g., a foosball table, colorful hats from previous gigs of the KMi rock band, a KMi-branded welcome pod at the main entrance). The objective was then to associate each category in this catalogue to a series of typical size features, represented in qualitative terms. To this aim, we isolated three features contributing to the size of objects, namely their (i) \textbf{front surface area} (i.e., the product of their width by their height), (ii) \textbf{depth} dimension, and (iii) \textbf{Aspect Ratio (AR)}, i.e., the ratio of their width to their height. With respect to the first dimension, we can characterise objects as \textit{extra-small}, \textit{small}, \textit{medium}, \textit{large} or \textit{extra-large} respectively. Secondly, objects can be categorised as \textit{flat}, \textit{thin}, \textit{thick}, or \textit{bulky}, based on their depth. Thirdly, we can further discriminate objects based on whether they are \textit{taller than wide} (\textit{ttw}), \textit{wider than tall} (\textit{wtt}), or \textit{equivalent} (\textit{eq}), i.e., of AR close to 1. If the first two qualitative dimensions were plotted on a cartesian plane, a series of quadrants would emerge, as illustrated in Figure 1. Then, the AR can help to further separate the clusters of objects belonging to the same quadrant. For instance, doors and desks both belong to the extra-large and bulky group but doors, contrarily to desks, are usually taller than wide.

Having defined the cartesian plane of Figure 1, we can manually allocate the KMi objects to each quadrant and further sort the objects lying in the same quadrant. Sorting the objects manually ensures more reliable results than if the same information was retrieved automatically, especially given the paucity of resources encoding the relative size of objects \cite{chiatti_towards_2020,bagherinezhad_are_2016}. 

\begin{figure}
  \centering
  \includegraphics[scale=0.2]{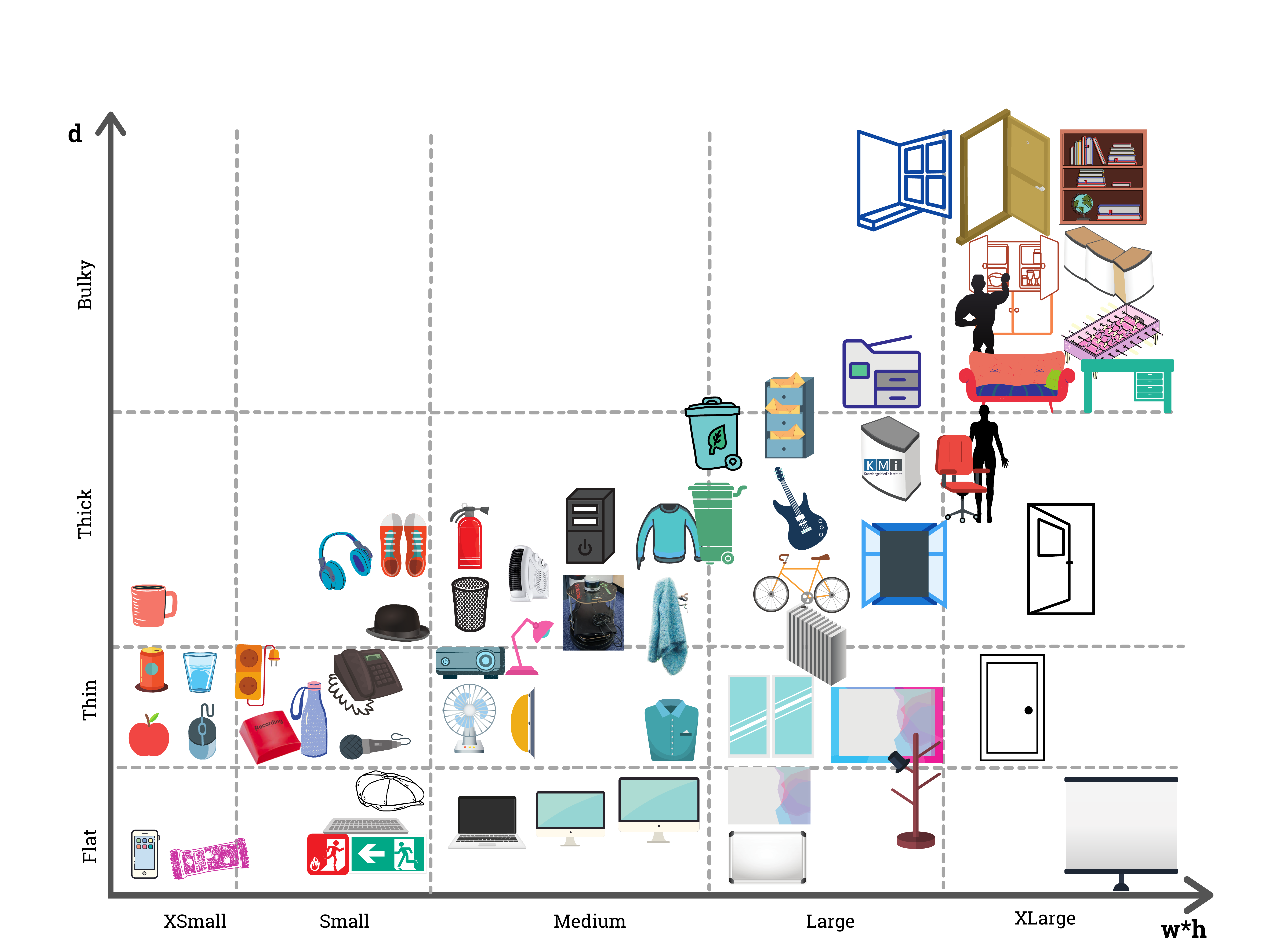}
  \caption{Examples of object sorting across two dimensions: (i) the area of the front surface (the product of the width w by the height h), on the x axis; and (ii) the depth value d, on the y axis.}
\end{figure}

Moreover, in the proposed representation, membership of each bin is mutually non-exclusive. Thus, with this representation, even classes which are extremely variable with respect to size, such as carton boxes and power cords, can be modelled. Indeed, boxes come in all sizes and power cords come in different lengths. Moreover, a box might lay completely flat, or appear bulkier, once assembled. Similarly, power cords, which are typically thinner than other pieces of IT equipment, might appear rolled up or tangled.

\subsection{Hybrid Reasoning Architecture}

We propose a modular approach to combining knowledge-based reasoning with Machine Learning for object recognition. In the proposed workflow, the knowledge of the qualitative size of objects (Section 3.1) is integrated in post-processing, after generating the ML-based object predictions. The proposed hybrid architecture is outlined in Figure 2 and consists of a series of sub-components, organised as follows. 

\begin{figure}
  \centering
  \includegraphics[width=\linewidth,trim=0 250 0 0, clip]{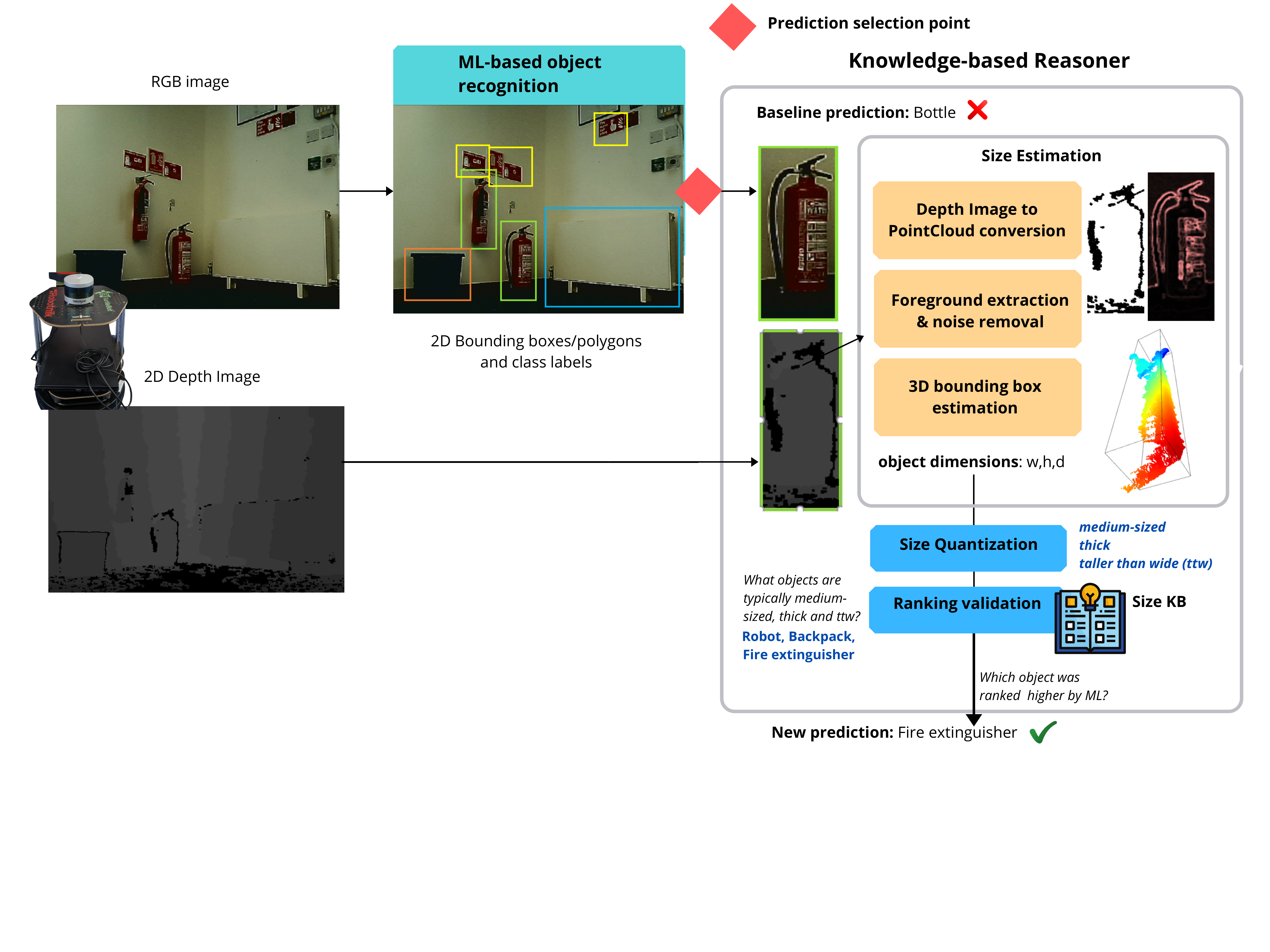}
  \caption{The proposed architecture for hybrid object recognition. The knowledge-based reasoning module, which is aware of the typical size features of objects, can be modularly queried to validate the ML-based predictions.}
\end{figure}

\textbf{ML-based object recognition.} We rely on the state-of-the-art, ML-based object recognition methods of \cite{zeng2018robotic}, to classify a set of pre-segmented object regions. Specifically, we classify objects by similarity to a reference image set, through a multi-branch Network. In this deep metric learning setting, predictions are ranked by increasing Euclidean (or L2) distance between each target embedding and the reference embedding set. Nevertheless, this configuration can be easily replaced by any other algorithm that provides, for each detected object, (i) a set of class predictions with an associated measure of confidence (whether similarity-based or probability-based) and (ii) the segmented image region enclosing the object. 

\textbf{Prediction selection.} This checkpoint is conceived for assessing whether a ML-based prediction needs to be corrected or not. At the time of writing, we achieved good results simply by retaining those predictions which the ML algorithm is most confident about, and by running the remaining predictions through the size-based reasoner. Specifically, we avoid the knowledge-based reasoning steps if the top-1 class in the ML ranking: (i) has a ranking score smaller than $\epsilon$ (i.e., the test image embedding lies near one of the reference embeddings, in terms of L2 distance); and also (ii) appears at least $i$ times in the top-K ranking. However, in Section 5, we also evaluate performance: (i) in the best case scenario where the ground truth labels are known and we know exactly which predictions to select for correction; and (ii) in the case where all ML-based predictions are passed to the size-based reasoner, without a pre-selection. 

\textbf{Size estimation.} At this stage, the input depth image corresponding to each RGB scene is first converted to a 3D PointCloud representation. Then, statistical outliers are filtered out to reduce the impact of noisy points and extract the dense 3D region which best approximates the volume of the observed object. Specifically, all points which lie farther away than two standard deviations ($2\sigma$) from their $n$ nearest neighbours are discarded. Because this outlier removal step is computationally expensive, especially for large 3D regions, we first sample each input PointCloud down so that 1 every $\chi$ points is retained. Then, the Convex Hull algorithm is used to approximate the 3D box bounding the object region. Lastly, the x,y,z dimensions of the 3D bounding box are computed. Since the orientation of the object is not known a priori, we cannot unequivocally map any of the estimated dimensions to the object’s real width and height. However, we can assume the object’s depth to be the minimum of the returned x,y,z dimensions, due to the way depth measurements are collected through the sensor. Indeed, since we do not apply any 3D reconstruction mechanisms, we can expect that the measured depth underestimates the real depth occupied by the object.

\textbf{Size quantization.} The three dimensions obtained at the previous step are here expressed in qualitative terms. First, the two dimensions which were not marked as depth are multiplied together, giving a proxy of the object’s surface area. The object is then categorised as extra-small, small, medium, large or extra-large, based on a set of cutoff thresholds $T$. Second, with respect to the estimated depth dimension, the object is categorised both as flat /non-flat (based on a threshold $\lambda_0$ ), and as flat, thin, thick, or bulky (based on a second set of thresholds $\Lambda$, where $\lambda_0 \in \Lambda$). Third, hypotheses are made about whether the object is taller than wide (ttw), wider than tall (wtt), or equivalent (eq), based on a cutoff $\omega_0 \in \Omega$. It would be unfeasible to predict the object’s Aspect Ratio from the estimated 3D dimensions, without knowing its current orientation. Therefore, we estimate the object’s AR based on the width (w) and height (h) of the 2D box bounding the object region, as follows:
\begin{equation}
AR = \begin{cases}
ttw &\text{if $h \geq w \land \frac{h}{w} \geq \omega_0$}\\
wwt &\text{if $h<w \land \frac{w}{h} \geq \omega_0$}\\
eq &\text{otherwise}
\end{cases}
\end{equation}

\textbf{Ranking validation.} The qualitative features returned by the size quantization module are then matched against the background KB introduced in Section 3.1, to identify the set of object categories which are plausible candidates for the observed object, based on their size. In Section 4, we test the effects of combining different size features (i.e., the area surface, thinness, and AR) to generate this candidate set. Ultimately, only those object classes in the original ML ranking which were validated as plausible with respect to the size are retained. 

\section{Experiments}
\subsection{Data Preparation}
\subsubsection{KMi RGB Reference Set}
For training purposes, RGB images depicting 60 object classes commonly found in KMi were collected through a Turtlebot mounting an Orbbec Astra Pro RGB-Depth (RGB-D) monocular camera. Each object was captured against a neutral background and opportunely cropped to control for the presence of clutter and occluded parts. Coherently to our prior experimental setup \cite{chiatti2020task}, we collected 10 images per class, with an 80\%-20\% training-validation split. Specifically, 5 images per class were used as anchor examples and paired up with their most similar image among the remaining 5 examples in that class (i.e., the multi-anchor switch strategy in  \cite{zeng2018robotic}). Image similarity was assessed by: (i) extracting features from the pre-trained ResNet50 module, (ii) L2-normalising the resulting image embeddings, and (iii) computing their cosine similarity. Then, we also matched each anchor with the most similar image belonging to a different class. In this way, instead of picking negative examples randomly, i.e., the protocol followed in \cite{zeng2018robotic}, we focused on triplets which are relatively harder to disambiguate.

\subsubsection{KMi RGB-D Test Set}
For testing purposes, additional RGB images and depth measurements of the KMi office environment were collected, during one of HanS’ monitoring routines. As a result, the class cardinalities in this set reflect the natural, relative occurrence of objects in the observed domain. For instance, HanS is more likely to spot fire extinguishers and radiators than printers, on its scouting route. These data were initially recorded as Robot Operating System (ROS) bag files, with both RGB and depth messages logged at a 640x480 resolution (i.e., the maximum resolution allowed by the depth sensor). 622 clear RGB frames, where the robot camera was still, were manually selected from the original recording.

Because the recording of RGB and depth messages is asynchronous, each RGB frame had to be matched to its nearest depth image, within a time window of +/- $\mu$. Choosing a higher value for $\mu$ increases the number of scenes for which a depth match is available, but also increases the chances that a certain scene is matched with the wrong depth measurements (e.g., because the robot has moved in the meantime). In our trials, we found that setting $\mu$ to 0.2 seconds offered a good compromise. With this setup, a depth match was found for 509 (~82\%) of the 622 original frames. After a visual inspection, only 2 (~0.4\%) of the 509 depth matches were identified as inaccurate and discarded. This RGB-D set was further pruned to exclude identical images, where neither the robot’s viewpoint nor the object arrangement had changed.

We annotated the remaining 213 images with respect to 60 reference object categories, through the Make Sense open-source annotation tool (https://www.makesense.ai/). Besides labelling each object, we also annotated its bounding rectangular or polygonal region, depending on its 2D shape. Resorting to polygonal masks to segment non-rectangular or partially occluded objects allowed us to produce higher quality annotations and to mitigate the effects of marginal noise. These annotations were used to crop the original RGB images and store each object region as a separate test example. The same annotated regions were then reused to crop the depth image linked to each RGB frame. In this case, the output crop was stored as a 16bit grayscale PNG, where non-zero pixels encode the depth values in millimeters. Upon analysing the generated depth crop, we identified 19 object regions which did not enclose any depth measurement. This can happen, for instance, when the object falls outside the range of the depth sensor (i.e., 60 cm – 8 m). To fairly compare the performance of the size-based reasoner (which relies on depth data) against the ML baseline, we discarded the empty depth crops from our test set, leaving us with a total 1414 object regions. 

\subsection{Ablation Study}
In what follows, we list the different methods under evaluation and illustrate the changes introduced before each performance assessment.

\textbf{Baseline Nearest Neighbour (NN).} In this pipeline, feature vectors are extracted from a ResNet50 module pre-trained on ImageNet \cite{he_deep_2016}, without re-training on the KMi RGB reference set. Namely, a 2048-dimensional embedding is extracted for each object region in the KMi RGB-D test set and matched to its nearest embedding in the KMi RGB reference set, in terms of L2 distance. This baseline provides us with a lower bound for the ML performance, before fine-tuning on the domain of interest.

\textbf{N-net} is the multi-branch Network which ensured the top performance on novel object classes, i.e., classes unseen at training time, in \cite{zeng2018robotic}. In this configuration, each CNN branch is a ResNet50 module pre-trained on ImageNet. Image triplets are formed following the methodology discussed in Section 4.1.1. At training time, hyperparameters are updated so that the Triplet Loss is minimised. In this way, the Network optimisation is directed towards minimising the L2 distance between matching pairs, while also maximising the L2 distance between dissimilar pairs. At inference time, object regions in the KMi RGB-D test set are classified based on their nearest object in the KMi RGB reference set, as in the case of the baseline NN pipeline.

\textbf{K-net} is the multi-branch Network which led to the top performance on known object classes, i.e., classes seen at training time, in \cite{zeng2018robotic}. K-net is a variation of N-net where a second loss component is added to the Triplet Loss. This auxiliary component of the loss derives from applying a SoftMax function to the last fully connected layer of the Network, to classify the input positive example over M classes. 

\textbf{Hybrid (area).} This configuration follows the hybrid architecture introduced in Section 3.2. However, only the object’s surface area is used as size feature to validate the ML predictions.

\textbf{Hybrid (area + flat).} With this ablation, we evaluate the effects of introducing a second feature to represent the size of objects. Specifically, here we consider not only the qualitative surface area of each object, but also whether they are flat or non-flat, based on their estimated depth. Then, the ML-based predictions are validated based on the set of object classes which both lie within the same area range and are also equally flat (or non-flat).

\textbf{Hybrid (area + thin)} is equivalent to the previous configuration, except the depth of objects is represented on a four-class scale: i.e., as flat, thin, thick, or bulky. The purpose of this ablation is testing the utility of introducing more granular bins in the y axis of Figure 1.

\textbf{Hybrid (area + flat + AR).} In this pipeline, we also integrate the qualitative Aspect Ratio (i.e., taller than wide, wider than tall, or equivalent) as a third knowledge property representing size. This setup is used for testing whether the qualitative AR can help further separating the object clusters exemplified in Figure 1.

\textbf{Hybrid (area + thin + AR) }follows the same configuration as the latter pipeline, except the object’s depth is categorised as as flat, thin, thick, or bulky. 

\subsection{Implementation Details}

\textbf{ML setup.} All the ML models illustrated in the previous Section were implemented in PyTorch \cite{paszke_pytorch_2019}. Images in the KMi RGB reference set were resized to 224 × 224 frames and normalised to the same distribution as the ImageNet-1000 dataset, which was used for pre-training the ResNet50 modules in each CNN branch. The tested ablations were fine-tuned through an Adabound optimizer \cite{luo_adaptive_2018}, over minibatches of 16 image triplets, with a learning rate set to start at 0.0001 and to trigger switching to Stochastic Gradient Descend optimization at 0.01. Parameters were updated for up to 1500 epochs, with an early stopping whenever the validation loss had not decreased for more than 100 epochs.

\noindent \textbf{Knowledge-based reasoning parameters. }We relied on the Python Open3D library \cite{zhou2018open3d} to process the PointCloud data and estimate the bounding 3D rectangles. During our trials we achieved the best results with threshold values set as follows. With respect to prediction selection, $\epsilon$ was set to a distance of 0.04 and $i$ to 3 predictions, within a top-5 ranking, i.e., with $K=5$. The three cutoff sets in the size quantization module were defined so that $T=$\{0.007,0.05,0.35,0.79\} (with thresholds expressed in squared meters), $\Lambda=$\{0.1,0.2,0.4\} (in meters) and $\Omega$=\{1.4 times\}. 

As an additional output of this work, we have also publicly released the RGB-D image set, annotated knowledge properties, and models used in these experiments: \url{https://github.com/kmi-robots/object_reasoner}. 

\section{Results and Discussion}
Performance on the KMi RGB-D test set was measured (i) in terms of the cross-class Accuracy, Precision (P), Recall (R) and F1 score of predictions in the top-1 result of the ranking; as well as (ii) based on the top-5 predictions in the ranking, in terms of mean Precision (P@5), mean normalised Discounted Cumulatve Gain (nDCG@5) and hit ratio (i.e., the ratio of the number of times the correct prediction appeared in the top-5 ranking to the total number of predictions). Specifically, the P, R and F1 metrics were aggregated across classes before and after weighing the averages by class support (i.e., based on the number of ground truth instances in each class). Measures@5 are unweighted. When comparing the different methods under evaluation, we prioritise improvements on the weighted F1 score, which accounts for the naturally imbalanced occurrence of classes in the test set. Moreover, at comparable top-1 results, we favour methods which provide higher quality top-5 rankings. Indeed, if the correct class was not ranked first but still appeared in the top-5 ranking, it would be easier for an additional reasoner (or human oracle) to correct the prediction.

\begin{table}
  \caption{ML baseline results on the KMi RGB-D test set.}
  \label{tab:ML}
  \begin{tabular}{l*{10}{l}}
    \toprule
    &  &\multicolumn{3}{c}{Top-1 unweighted} & \multicolumn{3}{c}{Top-1 weighted} & \multicolumn{3}{c}{Top-5 results unweighted} \\
    \hline
    Method & Top-1 Acc. & P & R & F1 & P & R & F1 & P@5 & nDCG@5 & Hit ratio \\ 
    \midrule
    Baseline NN & .33 &.36&.33&.25&.63&.33&.36&.23&.25&.54\\
    N-net & .45 &.34&\textbf{.40}&.31&.62&.45&.47&.33&.36&.63\\
    K-net & \textbf{.48} &\textbf{.39}&\textbf{.40}&\textbf{.34}&\textbf{.68}&\textbf{.48}&\textbf{.50}&\textbf{.38}&\textbf{.41}&\textbf{.65}\\
  \bottomrule
\end{tabular}
\end{table}

\begin{table}
  \caption{Hybrid reasoning results, when correcting only the wrong predictions.}
  \label{tab:best}
  \begin{tabular}{l*{10}{l}}
    \toprule
    &  &\multicolumn{3}{c}{Top-1 unweighted} & \multicolumn{3}{c}{Top-1 weighted} & \multicolumn{3}{c}{Top-5 results unweighted} \\
    \hline
    Method & Top-1 Acc. & P & R & F1 & P & R & F1 & P@5 & nDCG@5 & Hit ratio \\ 
    \midrule
    Hybrid (area) & .60&.52&.50&.47&.71&.60&.61&.43&.47&.72\\
    Hybrid (area+flat) & .60&.55&.50&.48&.72&.60&.62&.44&.47&.72\\
    Hybrid (area+thin) & .61&.55&.50&.48&.71&.61&.63&.44&.47&.71\\
    Hybrid (area+flat+AR)& \textbf{.62}&.59&.50&.49&\textbf{.76}&\textbf{.62}&\textbf{.65}&\textbf{.45}&\textbf{.49}&\textbf{.75}\\
    Hybrid (area+thin+AR) & \textbf{.62}&\textbf{.62}&\textbf{.51}&\textbf{.52}&.74&\textbf{.62}&\textbf{.65}&\textbf{.45}&.48&.74\\
  \bottomrule
\end{tabular}
\end{table}

First, we evaluated all methods which are purely based on Machine Learning, i.e., before any background knowledge about the typical size of objects is integrated. The results of this first assessment are reported in Table 1. K-net is the ML baseline which led to the top performance, across all evaluation metrics. Therefore, we considered the K-net predictions as a baseline, for testing all the hybrid configurations.

Because the knowledge-based reasoner relies on different sub-modules (i.e., prediction selection, size estimation and ranking validation), and each one of these modules is likely to propagate its own errors, we initially tested performance assuming that the ground truth predictions are known and that we can accurately discern which ML predictions need to be corrected. Although unrealistic, this best-case scenario provides us with an upper bound for the reasoner’s performance and aids the analysis of errors. As shown in Table 2, simply integrating knowledge about the qualitative surface area of objects already ensured a significant performance improvement, with a \textbf{13\%} increase of the unweighted F1 score and a \textbf{11\%} increase of the weighted F1 score. Overall, the best performance, both in terms of top-1 predictions as well as in terms of top-5 rankings, was achieved through the two hybrid configurations which included all the qualitative size features (i.e., surface area, thinness and AR). In particular, the unweighted F1 score increased up to \textbf{18\%} and the weighted F1 score up to \textbf{15\%}. Hence, the margin for improvement when complementing ML with size-based reasoning is significant. These results confirm the hypothesis laid in \cite{chiatti_towards_2020}: the capability to compare objects by size and the access to background knowledge representing size play a crucial role in object categorisation. Notably, there is no significant difference between the results obtained when representing depth in binomial terms (i.e., as either flat or non-flat), as opposed to when more fine-grained categories are used (i.e., flat, thin, thick or bulky). Thus, we can hypothesise that the costs (and potential inaccuracies) associated with formalising additional priors for the objects’ depth are not justified by a sufficient performance gain. 

\begin{table}
  \caption{Hybrid reasoning results, when correcting all predictions.}
  \label{tab:worst}
  \begin{tabular}{l*{10}{l}}
    \toprule
    &  &\multicolumn{3}{c}{Top-1 unweighted} & \multicolumn{3}{c}{Top-1 weighted} & \multicolumn{3}{c}{Top-5 results unweighted} \\
    \hline
    Method & Top-1 Acc. & P & R & F1 & P & R & F1 & P@5 & nDCG@5 & Hit ratio \\ 
    \midrule
    Hybrid (area)& \textbf{.49} &\textbf{.34}&\textbf{.33}&\textbf{.29}&\textbf{.59}&\textbf{.49}&\textbf{.50}&\textbf{.37}&\textbf{.40}&\textbf{.62}\\
    Hybrid (area+flat)& .47 &.33&.30&.27&.58&.47&.48&.37&.40&.60\\
    Hybrid (area+thin)& .44 &.30&.25&.23&.53&.44&.44&.34&.37&.54\\
    Hybrid (area+flat+AR)&.42 &.29&.22&.21&.58&.42&.44&.32&.35&.56 \\
    Hybrid (area+thin+AR)& .39 &.28&.20&.19&.53&.39&.40&.30&.32&.50\\
  \bottomrule
\end{tabular}
\end{table}

\begin{table}
  \caption{Hybrid reasoning results, when correcting only an automatically selected subset of predictions.}
  \label{tab:final}
  \begin{tabular}{l*{10}{l}}
    \toprule
    &  &\multicolumn{3}{c}{Top-1 unweighted} & \multicolumn{3}{c}{Top-1 weighted} & \multicolumn{3}{c}{Top-5 results unweighted} \\
    \hline
    Method & Top-1 Acc. & P & R & F1 & P & R & F1 & P@5 & nDCG@5 & Hit ratio \\ 
    \midrule
    Hybrid (area) & \textbf{.55}&.42&\textbf{.41}&\textbf{.38}&.67&\textbf{.55}&\textbf{.57}&\textbf{.43}&\textbf{.46}&\textbf{.69}\\
    Hybrid (area+flat) & .54&\textbf{.43}&.39&.37&.67&.54&.56&\textbf{.43}&\textbf{.46}&.68\\
    Hybrid (area+thin) & .52&.39&.37&.35&.62&.52&.54&.42&.44&.64\\
    Hybrid (area+flat+AR) & .53&\textbf{.43}&.37&.36&\textbf{.69}&.53&.56&\textbf{.43}&.45&.68\\
    Hybrid (area+thin+AR) & .51&\textbf{.43}&.36&.36&.64&.51&.54&.41&.43&.63\\
  \bottomrule
\end{tabular}
\end{table}

Then, we ran all the predictions returned by the K-net method through the hybrid reasoning pipelines, without discriminating between correct and incorrect ML predictions (Table 3). This experiment is designed to evaluate the performance of the size-based reasoner in the worst-case scenario, i.e., when no assumption can be made about the ML algorithm confidence. As expected, the obtained results are comparable to or inferior to those achieved through a purely ML-based pipeline. These results indicate that applying the size priors indiscriminately led to inaccurate corrections of objects that the ML algorithm had correctly recognised. This issue is further emphasized the more size features (or knowledge priors) are introduced. As shown in Table 3, including other characteristics besides the object’s surface area causes a performance drop.

Therefore, to capitalise on the latent performance gains highlighted in Table 2, the ML and knowledge based outcomes need to be opportunely leveraged. To this aim, we introduced a meta-reasoning checkpoint (i.e., the prediction selection module described in Section 3.2) and automatically selected a subset of ML predictions to feed to the knowledge-based reasoner. The results of this last evaluation setup are summarised in Table 4. The tested hybrid configurations allowed to reach a compromise between the lower performance bound of Table 3 and the upper performance bound of Table 2. Specifically, the ML baseline was outperformed by up to \textbf{4\% }in terms of unweighted F1 and by up to \textbf{7\% } in terms of weighted F1. Moreover, introducing knowledge about the object’s surface positively impacted the quality of the top-5 ranking: the mean P@5 and nDCG@5 both increased by \textbf{5\%}, and the hit ratio by \textbf{4\%}. Similarly to the results in Table 3, the qualitative surface area is the feature which led to the most consistent results across the different evaluation metrics. In other words, integrating additional knowledge beyond that first feature only led to comparable results, or even degraded the performance (i.e., in the case where a four-class scale instead of a binary one is used for the object’s depth). Hence, in the experimental scenario of this paper, a size representation as minimalistic as indicating whether the object exposes an extra-small, small, medium, large, or extra-large front surface area is sufficient to ensure a significant boost in performance. 

\section{Conclusion and Future Work}
In this paper we demonstrated that augmenting an object recognition pipeline based on Machine Learning with an effective representation of object sizes and a size reasoner can significantly improve the quality of predictions, as hypothesised in our prior work \cite{chiatti_towards_2020}. These results are particularly promising, because they were achieved on image regions collected by a robot in its natural environment, i.e., in a more challenging experimental setup than the standard classification of benchmark image collections. In the proposed approach, we relied on a novel representation of the size of objects. Differently from prior knowledge representations, here we modelled size across three dimensions (the object’s front surface area, depth and aspect ratio), to further separate the object clusters. Moreover, we allowed for annotating each object class with different size attributes, to adequately capture the size variability within each class.

The experiments presented in this paper also highlighted a series of directions of improvement, informing our future work. First, when estimating the object size from depth data we had to deal with hardware constraints: (i) objects falling outside the range of the depth sensors were excluded; (ii) highly reflective, absorptive or transparent materials (e.g., shiny metals, glass) altered the depth measurements. As such, access to a more advanced depth sensor would further improve the performance. Second, if the object was only partially visible in the original image, the estimated measurements (albeit accurate) would fail to represent the real object’s size. Thus, incorporating the capability of moving towards the target object to refine the prediction through repeated measurements (i.e., Active Vision) is likely to benefit performance. Third, in the proposed method, we finetuned parameters through repeated trials. To increase the efficiency and scalability of our solution across different application scenarios, we will explore alternative solutions to (partially) automate this parameter tuning routine.

Naturally, the relevance of background knowledge and knowledge-based reasoning for enabling Visual Intelligence spans way beyond the capability to reason about the typical size of objects. In \cite{chiatti_towards_2020}, we have identified several other reasoners (e.g., spatial, compositional, motion-aware) which may enhance the robustness of state-of-the-art Machine Learning methods. Thus, in our future work, we will evaluate the performance impacts of integrating: (i) additional knowledge-based components, (ii) multiple sources of background knowledge, as well as (iii) effective meta-reasoning strategies, to reconcile the outcomes of different reasoners.

\bibliography{main}


\end{document}